# The Design and Realization of Multi-agent Obstacle Avoidance based on Reinforcement Learning

Enyu Zhao[a,b], Chanjuan Liu[a*], Houfu Su[a], and Yang Liu[a]

[a] *Department of Computer Science and Technology, Dalian University of Technology, Dalian 116024, China*

[b] *University of Southern California Viterbi Department of Computer Science, 941 Bloom Walk, Los Angeles, CA 90089*

*Abstract*— Intelligence agents and multi-agent systems play important roles in scenes like the control system of grouped drones, and multi-agent navigation and obstacle avoidance which is the foundational function of advanced application has great importance. In multi-agent navigation and obstacle avoidance tasks, the decision-making interactions and dynamic changes of agents are difficult for traditional route planning algorithms or reinforcement learning algorithms with the increased complexity of the environment.

The classical multi-agent reinforcement learning algorithm, Multi-agent deep deterministic policy gradient(MADDPG), solved precedent algorithms' problems of having unstationary training process and unable to deal with environment randomness. However, MADDPG ignored the temporal message hidden beneath agents' interaction with the environment. Besides, due to its CTDE technique which let each agent's critic network to calculate over all agents' action and the whole environment information, it lacks ability to scale to larger amount of agents.

To deal with MADDPG's ignorance of the temporal information of the data, this article proposes a new algorithm called MADDPG-LSTMactor, which combines MADDPG with Long short term memory (LSTM). By using agent's observations of continuous timesteps as the input of its policy network, it allows the LSTM layer to process the hidden temporal message. Experiment's result demonstrated that this algorithm had better performance in scenarios where the amount of agents is small. Besides, to solve MADDPG's drawback of not being efficient in scenarios where agents are too many, this article puts forward a light-weight MADDPG (MADDPG-L) algorithm, which simplifies the input of critic network. The result of experiments showed that this algorithm had better performance than MADDPG when the amount of agents was large.

## I. Introduction

Intelligent agents have played important roles in daily life such as robots in factories or drones used in warfare. With the increased complexity of the environment, single-agent system's performance become unsatisfying. Thus multi-agent system drew people's attention with its multi-agent structure which enables it to finish complicated tasks in challenging environment by letting agents to communicate and cooperate with each other. Multi-agent system was applied in a range of regions like traffic regulation, computer games and robotic. The ability to navigation and obstacle avoidance is undoubtedly the essential feature of agents in multi-agent system as the wide utilization of mobile multi-agent system in areas like autonomous driving cars and robots in the storehouse.

At present, the development of autonomous navigation technology of agent system can be roughly divided into three directions according to the knowability of map data: method based on static path planning, method based on real-time map creation and positioning, and method based on reinforcement learning algorithm.

The navigation algorithm based on static path planning firstly models the mobile robot's activity scene, establishes the global map of the scene, and stores the map data of the scene in the robot's hardware device. The mainly used algorithms [4-7] are probabilistic roadmap method [1], rapid-exploration random tree method [2] and artificial potential field method [3]. However, the navigation algorithm based on static path planning can only be applied to the scene where the map is known.

The technology of simultaneous localization and mapping (SLAM) was proposed by Smith [8-9] in 1986. Many researchers [10-11] have made a variety of optimization attempts on map creation, positioning and path planning. However, due to the dynamic changes of environment and agents, the acquisition of sensor data that slam relies on is faced with uncertainty and complexity.

As the obstacle avoidance ability is in the subset of autonomous navigation, the researches done in autonomous navigation also included exploration in obstacle avoidance. However, the traditional route planning algorithms like A* [12] exhibited problems such as unable to adapt to dynamic environments efficiently, surging time complexity with the increase of agents. With the development made in deep learning, deep reinforcement learning was introduced in to the multi-agent system. Multi-agent reinforcement learning became a solution to the obstacle avoidance and navigation problem.

Preliminary design of multi-agent reinforcement learning directly makes each agent run a single reinforcement learning algorithm. IQL [13] is an example, each agent in this algorithm executes an individual Q-learning [14] algorithm. Another example is IDDPG with each agent run an individual DDPG [15] algorithm. Because agent can't have access to other agents' action or observation, it can't tell whether the change in environment is due to the action of other agents or just the environment's stochasticity. The instability in the environment, from the agent's perspective, will cause the agent fail to converge [15].

Multi-Agent Deep Deterministic Policy Gradient (MADDPG) [16] brought forward the Centralised Training and Decentralized Execution (CTDE) training method as agents can access other agents' action and observation during training phase, even the global observation of the whole environment. The CTDE training fashion can enable agent to comprehend the environment even with other agents' exploration to ensure the stability of the training process and leads to a higher performance. However, it's still not perfect as centralized training requires each agent to consider others' action and observation in the training period, so when the number of agents increased the joint action space will grow exponentially. To solve this problem, Value Decomposition Network (VDN) [17] adds all agents' local critic's output to gain the global action-value and QMIX [18] passes all agents' local action-value to the mixer network to estimate the value of the joint action of all agents. As Q-learning, the basic reinforcement learning algorithm of QMIX is only for discrete action space, and Factorized Multi-agent Actor-Critic(FACMAC) [19] was put forward as a subsidy for continuous action space.

As in the agent navigation and obstacle avoidance task, each agent's observation and action follows a continuous time series pattern, Long Short-Term Memory(LSTM) [20] and transformer [21] can be utilized in the reinforcement learning process to increase performance and stability. However, its application in Multi-agent reinforcement learning area still await exploration.

In this work, we propose two algorithms, viz., MADDPG-LSTMactor and MADDPG-L to solve the obstacle avoidance and navigation problem for multi-agent systems, especially focusing on whether the process of time series can improve algorithm's performance and how to enable the algorithm to scale to larger amount of agents. The training and execution of the algorithms and the underlying algorithms: IDDPG, MADDPG, FACMAC, were together tested in several virtual environments. The results suggest that the MADDPG-LSTMactor algorithm performs better than the underlying MADDPG as well as other baseline algorithms. When the number of agents increases, the MADDPG-L algorithm demonstrates the best performance. Therefore, the proposed algorithms can be used as a basic algorithm for multi-agent obstacle avoidance navigation tasks in the future.

The paper is organized as follows: Section 2 introduces the preliminaries for later use; In Section 3, we introduce the idea of the algorithms LSTM-MADDPG and MADDPG-L; Section 4 discusses the experimental environment and the comparative analysis of the experimental results; Section 5 summarizes the paper with possible research directions in the future.

## II. BACKGROUND

### A. DDPG

In Deterministic Policy Gradient (DPG), the deterministic policy $\mu$ will output a certain action $u$ in given state $s$, $u = \mu_\theta(s)$ in contrary to the Stochastic policy which gives the probability of different actions. This simplifies the calculation of policy gradient by only executing integration on state rather than both on state and action. The policy gradient is calculated in the way shown in Equation (1).

$$\nabla_\theta J(\theta) = \mathbb{E}_{s \sim \mathcal{D}}[\nabla_\theta \mu_\theta(u|s) \nabla_u Q^\mu(s,u)|_{u=\mu_\theta(s)}] \quad (1)$$

Deep Deterministic Policy Gradient (DDPG) [11] adapts neural network to serve as the policy network $\mu$ and evaluation function $Q^\mu$. And IDDPG is the simple extension of the DDPG in the multi-agent field by executing an independent DDPG algorithm on each agent.

### B. MADDPG

MADDPG is one of the classic algorithms for multi-agent reinforcement learning. It adopts the training method of Centralized Training Distribution Execution (CTDE) and extends the Deep Deterministic Policy Gradient (DDPG) [22] algorithm to the multi-agent domain. It uses offline learning to train agents that perform actions within a continuous action space. In the MADDPG algorithm, each agent will be trained with the basic structure of actor-critic, learning the deterministic policy network $\mu_a$ and the action evaluation function $Q^{\mu_a}$ exclusive to the agent $a$, respectively. Since each agent's policy and action evaluation function belong only to the agent itself, this makes it applicable to totally competitive, totally-cooperative or mixed-setting environments.

In this article, agent a's policy network $\mu_a$ will be parameterized by $\theta_a$, that is $\mu_a(\tau_a; \theta_a)$. The joint policy of all agents will be noted as $\boldsymbol{\mu}$, $\boldsymbol{\mu} = \{\mu_a(\tau_a; \theta_a)\}_{a=1}^n$。Each agent's critic network is noted as $Q^{\mu_a}$, with $\phi_a$ as its parameters. The critic network will evaluate agent a's action by the state s of the environment and the joint action u of all agents, which is $Q^{\mu_a}(s, u_1, u_2 \ldots u_n; \phi_a)$。The critic network of agents $Q^{\mu_a}$ will be trained in time differential fashion, aiming to minimize loss $\mathcal{L}(\phi_a)$, as in Equation (2):

$$L(\phi_a) = E_D\left[\left(y^a - Q^{\mu_a}(s, u_1, u_2 \ldots u_n; \phi_a)\right)^2\right] \quad (2)$$

Where $y^a = r_a + \gamma Q^{\mu_a}(s', u'_1, u'_2 \ldots u'_n|_{u'_a=\mu_a(\tau_a;\theta_a^-)}; \phi_a^-)$。 $r_a$ is the reward gained by agent $a$'s interaction with the environment, $\phi_a^-$ is the parameter of target critic network, $(u'_1, u'_2, \ldots u'_n)$ represent the target action of all agents chosen by their target policy network with parameter $\theta_a^-$。The experience tuple stored in replay buffer $\mathcal{D}$ is $(s, s', u_1, u_2, \ldots u_n, r_1, r_2, \ldots r_n)$。

The policy gradient $J(\mu_a)$ of agent $a$ is calculated in Equation (3).

$$\nabla_{\theta_a} J(\mu_a) = \mathbb{E}_\mathcal{D}[\nabla_{\theta_a} \mu_a(\tau_a) \nabla_{u_a} Q^{\mu_a}(s, u_1, u_2, \ldots u_n)|_{u_a=\mu_a(\tau_a)}] \quad (3)$$

It needs to be mentioned that except the action of agent a trained currently, which is computed by agent a's policy network, other agents' action is just the corresponding action stored in replay buffer.

*C. VDN and QMIX*

Both Value Decomposition Network (VDN) and Monotonic Q-Value Decomposition Network (QMIX) are multi-agent reinforcement learning algorithms that are trained using centralized training distribution execution. Both algorithm tasks are to train a centralized critic. Unlike the MADDPG algorithm, the environment action evaluation function they train will have only one output that evaluates the global action-state value, which is $Q_{tot}^\pi$. This environmental action evaluation function, which evaluates the global action-state value, will be calculated in a decomposed fashion. The practice of the value decomposition network (VDN) is to add the environmental action evaluation functions of each agent, as shown in Equation (4):

$$Q_{tot}^\pi(\boldsymbol{\tau}, \boldsymbol{u}; \phi) = \sum_{a=1}^{n} Q_a^{\pi_a}(\tau_a, u_a; \phi_a) \quad (4)$$

It should be noted that the environmental action evaluation function $Q_a^{\pi_a}$ of each agent is only evaluated according to the local observation record and the action of the agent $a$, unlike the counterparts in MADDPG which consider the overall environment state and the actions of all other agents.

The monotonic Q-value decomposition network (QMIX) uses a monotonic continuous mixing function to calculate the global action-state evaluation function $Q_{tot}^\pi$, as shown in Equation (5):

$$Q_{tot}^\pi(\boldsymbol{\tau}, \boldsymbol{u}, s; \boldsymbol{\phi}, \psi) = f_\psi\left(s, Q_1^{\pi_1}(\tau_1, u_1; \phi_1), Q_2^{\pi_2}(\tau_2, u_2; \phi_2) \dots Q_n^{\pi_n}(\tau_n, u_n; \phi_n)\right) \quad (5)$$

The requirement for the monotonicity of the mixing function is to ensure that when the global action-state evaluation function $Q_{tot}^\pi$ is the largest, the actions selected by each agent are the same as those selected when the agent $Q_a^{\pi_a}$ is maximized to facilitate distribution. In practical tasks, the mixing function in QMIX is a neural network, which guarantees monotonicity by using non-negative weights. The monotonic QMIX algorithm still uses the time differential algorithm for training with the target network, and find the optimal solution by reducing the loss function $L(\phi, \psi)$, as shown in Equation (6):

$$L(\boldsymbol{\phi}, \psi) = E_D\left[\left(y^{tot} - Q_{tot}^\pi(\boldsymbol{\tau}, \boldsymbol{u}, s; \boldsymbol{\phi}, \psi)\right)^2\right] \quad (6)$$

Where $y^{tot}$ is:

$$y^{tot} = r + \gamma \max_{u'} Q_{tot}^\pi(\boldsymbol{\tau'}, \boldsymbol{u'}, s'; \boldsymbol{\phi}^-, \psi^-)$$

$\boldsymbol{\phi}^-$ is the set of network parameters for the target critic network of each agent, and $\psi^-$ is the target mixing network parameter.

*D. FACMAC*

Since VDN and QMIX adopt the basic idea of Q-learning, these two algorithms are only suitable for discrete action spaces. The factored multi-agent centralized policy gradient algorithm (FACtored Multi-Agent centralized policy gradient, FACMAC) can be used to deal with multi-agent reinforcement learning tasks in continuous action spaces.

As MADDPG trains the centralized action evaluation network $Q_{tot}^\mu$ in CTDE method, that is, to learn a centralized critic for each agent a to evaluate the overall environment and all agent actions. The action evaluation network $Q_a^{\mu_a}$ becomes more difficult to optimize as the number of agents increase and action space surge. Because with the increase of the number of agents and the action space of each agent in the multi-agent system, the dimension of the input vector $(s, u_1, u_2 \dots u_n)$ of the environmental action evaluation network $Q_a^{\mu_a}$ of each agent will increase exponentially. Therefore, the Factorized Multi-Agent Centralized Policy Gradient Algorithm (FACMAC) proposes the use of value factorisation to construct and train a centralized environmental action evaluation function $Q_{tot}^\mu$ that is easier to cope with the increase in the number of agents and the increase in the action space.

Since FACMAC adopts the Actor-Critic structure as the basic algorithm framework, each agent $a$ relys on the unique action $u_a$ calculated by its policy network $\mu_a$ according to the local action observation history $\tau_a$ when making action selection. Unlike VDN and QMIX, agent only depends on the action value given by the action evaluation function $Q_{tot}^\mu$ to choose action. This means when constructing the factored and centralized critic $Q_{tot}^\mu$, there is no need to impose monotonicity restrictions on the mixing function. This enables the FACMAC to more freely decompose the centralized environmental action evaluation function $Q_{tot}^\mu$ without considering its possible performance loss, so as to better deal with more complex surroundings.

The centralized action-value evaluation function of FACMAC is shown in Equation (7):

$$Q_{tot}^\mu(\boldsymbol{\tau}, \boldsymbol{u}, s; \boldsymbol{\phi}, \psi) = g_\psi\left(s, Q_1^{\mu_1}(\tau_1, u_1; \phi_1), Q_2^{\mu_2}(\tau_2, u_2; \phi_2) \dots Q_n^{\mu_n}(\tau_n, u_n; \phi_n)\right) \quad (7)$$

$\boldsymbol{\phi}$ and $\phi_a$ represent the parameters of the centralized action evaluation function $Q_{tot}^\mu$ and the local environmental action evaluation function $Q_a^{\mu_a}$ of each agent a, respectively. $g_\psi$ is a nonlinear mixing function, which is not restricted by monotonicity, and the parameter is $\psi$. The learning method is the same as the QMIX, the target network is used to learn the centralized action evaluation function. The loss function is expressed in Equation (8):

$$L(\boldsymbol{\phi}, \psi) = E_D\left[\left(y^{tot} - Q_{tot}^\mu(\boldsymbol{\tau}, \boldsymbol{u}, s; \boldsymbol{\phi}, \psi)\right)^2\right] \quad (8)$$

Where $y^{tot}$ is:

$$y^{tot} = r + \gamma Q_{tot}^\mu(\boldsymbol{\tau'}, \boldsymbol{\mu}(\boldsymbol{\tau'}; \boldsymbol{\theta}^-), s'; \boldsymbol{\phi}^-, \psi^-)$$

In the calculation of the policy gradient, FACMAC proposes a calculation method that computes the centralised policy gradient as in Equation (9).

$$\nabla_\theta J(\boldsymbol{\mu}) = \mathbb{E}_D[\nabla_\theta \boldsymbol{\mu} \nabla_{\boldsymbol{\mu}} \boldsymbol{Q}_{tot}^\mu(\tau, \mu_1(\tau_1), \mu_2(\tau_2) \dots \mu_n(\tau_n), s)] \quad (9)$$

Different from the gradient calculation method of MADDPG shown in Equation (2), in FACMAC, all agents' actions $u_a$ are calculated by the policy network $\mu_a$ of each agent $a$ according to its local action observation history $\tau_a$. While in MADDPG, only the action of the currently trained agent is calculated in the above way, other agents' action are sampled from the replay buffer.

## III. THE DESIGN OF ALGORITHMS

### A. MADDPG-LSTMactor

In multi-agent reinforcement learning tasks, agents interact with the environment using their own policies. Then they gain rewards resulting from their actions' influence on the environment. In this learning process, each agent's actions and observations as well as the states of the environment all demonstrate the features like time series. Due to LSTM's capability of dealing with time series data, we introduce this structure into MADDPG (which has been proved to be a successful algorithm in multi-agent reinforcement learning field) to see whether the performance in multi-agent obstacle avoidance and navigation tasks can be further enhanced.

As MADDPG algorithm adopts the CTDE (centralized training and Decentralized execution) training method, each agent's critic network estimates the action $u_a$ made by actor network based on the environment's state $s$ to guide the actor in training phase, whereas the actor network generates action both in training and execution phase using its observation history as input. So, it is obvious that comparing to the centralized critic network, the actor network deals with time series data more frequently and directly. Thus, we tried to introduce LSTM in the actor network directly first.

Normally, agents in MADDPG algorithm interact with the environment without training until there's enough data in replay buffer $\mathcal{D}$. Then, at each time step, a batch of experience tuples containing environment state , action $\boldsymbol{u} \equiv \{u_a\}_{a=1}^n$, reward $\boldsymbol{r} \equiv \{r_a\}_{a=1}^n$ will be sampled from the replay buffer $\mathcal{D}$ to train the critic network and the actor network. After the training phase, the updated actor network will continue to interact with the environment to generate new data. The environment will be reset periodically when the timestep becomes an integral multiple of preset hyper-parameter $episode - length$. However, such training process is unable to exploit the information hidden in the time series.

In this article, agents' policy network equipped with LSTM takes action $u_a$ on timestep $t$ based on a sequence of its local observation $\boldsymbol{\tau_a} \equiv \{\tau_a^{t-i}\}_{i=0}^l$ rather than simply the observation $\tau_a^t$ of timestep $t$. We set the sequence length as a hyper-parameter $squence - length = l$. The action-taking process of agent $a$ can be defined as Equation (10).

$$u_a = \mu_a(\{\tau_a^{t-i}\}_{i=0}^l, h^{t-1}, c^{t-1}; \theta_a) \qquad (10)$$

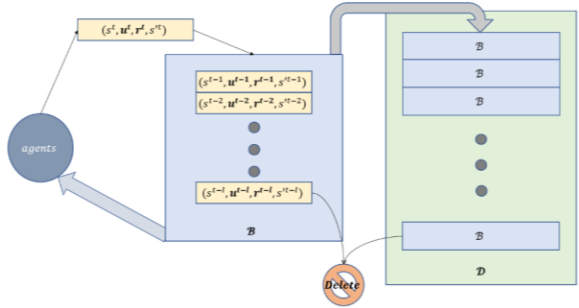

Figure 1. The illustration of MADDPG-LSTMactor

We realized this mechanism by creating and maintaining a queue with the length of $seq - length$ during both training and execution phase. The queue stores experience tuples and is named as sequence buffer $\mathcal{B}$. At each timestep, the actor network takes action $u_a$ based on the observation sequence stored in sequence buffer $\mathcal{B}$ and gains reward $r_a$. The experience tuple including environment state $s$, all agents' observation $\boldsymbol{\tau}$, all agents' action $\boldsymbol{u}$ and reward $\boldsymbol{r}$ will be stored into the sequence buffer. If the sequence buffer is full, the first experience tuple will be deleted and the newest one will be stored at the end of the queue. Meantime, the whole sequence buffer will be stored into the replay buffer $\mathcal{D}$ as an element. The sequence buffer will be emptied when the environment is reset. During training, instead of sampling experience tuples, the previously stored sequence buffer, which is a continuous sequence of experience tuples, will be sampled. The illustration figure is shown below.

In MADDPG-LSTMactor algorithm, the structure of critic network and how it works are same to the one in MADDPG. Thus, the way to optimize critic network and actor network remains unchanged.

### B. MADDPG-L

As a typical MARL algorithm adopts CTDE training method, MADDPG requires agents' critic network to consider every agent's observation and action in training phase to stabilize training as well as improve the algorithm's performance. However, such demand for extra information can be a problem when the amount of agents in the multi-agent system is too large, as it will lead to the explosion of action space. FACMAC can be a great solution to this question. By learning a single centralized but factor critic, which factors the joint action-value function $Q_{tot}$ into per-agent utilities $Q_a$ that are combined via a non-linear function. Each factored action-value utility, i.e., each agent's local critic network estimates the value of its action, which can be written as $Q_a(\tau_a, u_a; \phi_a)$. $\tau_a$ is the local observation of agent $a$, $u_a$ is its action and $\phi_a$ stands for the parameters of the critic network. Then the mixing function $g_\psi$ will take all agents' critic network's output and the environment's state $s$ to give the final joint action-value estimation for all agents. In this way, agent's local critic network only needs to calculate its own action and observation, regardless of the number of agents in the system. The mixing function helps to ensure that during training phase, all agents can get information about other agents and the environment.

However, algorithms factoring the centralized critic are not perfect. In the baseline test of MPE (Multi-agent Particle Environment) [23], the QMIX algorithm demonstrated the problem that it's hard to get away from the initial state. The result is shown in **Fig.** 2.

The horizontal axis is the timestep of training, while the vertical axis is the average return of the algorithm. It can be witnessed that the performance, which is indicated by its average return, doesn't make any practical improvement until the training timestep is more than 3 million. In this article's experiment, the FACMAC algorithm showed the same problem because the structure design and the idea of FACMAC are similar to QMIX. The result can be found in chapter 4.

We suggest that this problem could happen because there are too many networks to optimize in the training procedure.

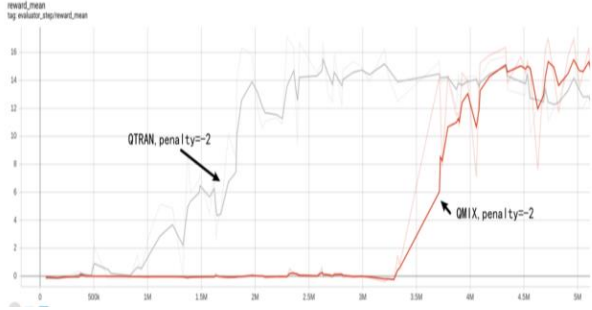

Figure 2. The performance of QMIX in MPE

As the centralized and final critic of FACMAC is created by the mixing function which mix the combination of all local critics, we need to optimize all of them while training each agent. In comparison to MADDPG, it only requires optimizing the agent's own critic when it's the agent's turn to be trained. So, it's harder for the FACMAC algorithm to make solid progress and this could lead to the cold start problem which worsen the situation. Because the algorithm is still in the primitive stage, most of its attempts can't bring any positive rewards, so its replay buffer is flooded with 'not successful' experience tuples for the training phase to sample. The reason why we made such suggestion comes from the way to calculate centralized critic in the experiment. When we are training an agent, we need to calculate the centralized critic, as the actor network need it to carry out gradient ascend. As the centralized critic requires other agents' critic result besides the agent $a$ which is currently trained, we can directly use other agents' critic to estimate the action-value of their own actions or use the critic network of agent $a$ to simulate other agents' critic network to produce the action-value. As in our experiment, all agents' critic networks share the same structure, which makes the simulation method possible. The advantage of using only the critic network of agent $a$ is it can reduce the amount of networks to improve, which presumably makes training easier. In our experiment, such method did improve the performance. However, the progress is still far from satisfaction.

In this article, we bring forward a light-weight MADDPG algorithm, which can better scale to larger amount of agents, we named it as MADDPG-L. The actor network remains the same as MADDPG, we set the policy of agent $a$ to be $\mu_a(\tau_a;\theta_a)$, and the action of agent $a$ to be $u_a = \mu_a(\tau_a,\theta_a)$. $\theta_a$ is the parameter of $\mu_a$. What distinguish it from MADDPG is the critic network of each agent. The notation of it is $Q_a^\mu(s, u_a; \phi_a)$, implying that the critic of each agent only considers that action of its own instead of all agents' action, $\phi_a$ is the parameter of $Q_a^\mu$. Meanwhile, the environment state $s$ helps the agent to get the global information it needs. This change helps to solve the problem of action space explosion as each agent only needs to calculate its own action regardless of the amount of agents in the multi-agent system. The illustration is shown in **Fig.** 3. The critic network is trained in temporal difference style, the loss function of it is shown in Equation (11).

$$\mathcal{L}(\phi_a) = \mathbb{E}_\mathcal{D}\left[\left(y^a - Q^{\mu_a}(s, u_a; \phi_a)\right)^2\right] \quad (11)$$

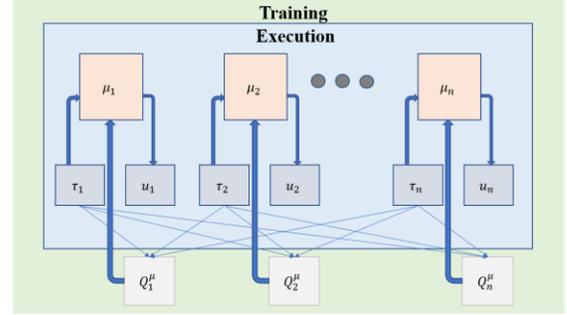

Figure 3. Illustration of MADDPG-L

where $y^a$ is:

$$y^a = r_a + \gamma Q^{\mu_a}(s', u'_a|_{u'_a = \mu_a(\tau_a;\theta_a^-)}; \phi_a^-)$$

$r_a$ is the reward of agent $a$'s interaction with the environment, $\phi_a^-$ indicates the parameter of target network, $u'_a$ is the action implemented by the target actor network with its parameter notified as $\theta_a^-$. The experience tuples stored in the replay buffer $\mathcal{D}$ has the structure as $(s, s', u_1, u_2, ... u_n, r_1, r_2, ... r_n)$.

The policy gradient of agent $a$'s actor network $J(\mu_a)$ is shown in Equation (12):

$$\nabla_{\theta_a} J(\mu_a) = \mathbb{E}_\mathcal{D}[\nabla_{\theta_a}\mu_a(\tau_a)\nabla_{u_a}Q^{\mu_a}(s, u_a)|_{u_a=\mu_a(\tau_a)}] \quad (12)$$

## IV. EXPERIMENT

### A. Simulation environment

The simulation environment is based on MPE environment, and the environmental parameters are listed in the Table 1.

### B. The experiment scenarios and results

The experiment scenarios are **Obstacle predator-prey**, **Spread**, **Tunnel** and **Simple Tunnel**. The **Obstacle Predator-Prey** acts as the initial test environment for all algorithms to find out their performance in obstacle-avoidance and dynamic target pursuing task. The **Spread** environment aims to test all algorithms' performance in avoiding obstacles that are relatively small and arriving at designated destinations. What's more, the Spread environment will have multiple versions including different number of agents to discover the algorithm's ability to scale. The **Tunnel** environment is meant for testing algorithm's ability in navigating through environments where obstacles are so huge that they form the contour. In other words, there will be some specific routes for the agents to take and algorithms should learn to find them. The **Simple Tunnel** environment is merely the Tunnel environment with some modifications to make it more realistic. Apart from the Simple Tunnel environment, the algorithms to be tested are IDDPG, MADDPG, MADDPG-LSTMactor, MADDPG-L, FACMAC. In the Simple Tunnel environment, agents' rewards are not equal and thus FACMAC is not tested as it requires such a condition.

TABLE I. EXPERIMENTAL ENVIRONMENT PARAMETERS

| Term | Version |
|---|---|
| CPU | Intel i9 9900k |
| Memory Size | 64GB |
| GPU | Colorful NVIDIA GTX1080(VRAM 8GB) |
| Operation System | Ubuntu 18.04 LTS |
| GPU Driver version | 495.46 |
| Python version | 3.8.13 |
| Pytorch version | 1.11.0 |
| CUDA version | 11.5 |

TABLE II. HYPERPARAMETERS OF THE EXPERIMENT

| Hyperparamters | Value | Meaning |
|---|---|---|
| max-episode-len | 100 | The length of each episode |
| time-steps | 2000000 | Total interaction steps |
| Num-adversaries | 1 (default) | Other agents besides those get trained |
| Lr-actor | 0.001 (default) | The learning rate of policy network |
| Lr-critic | 0.01 (default) | The learning rate of critic network |
| Epsilon | 0.1 | Parameter for $\epsilon\text{-}greedy$ action choosing policy |
| Noise-rate | 0.1 | The rate of noise added to action during training phase |
| Gamma | 0.95 | Time elapse coefficient |
| Batch-size | 256 (default) | Batch-size for experience tuples sampled for training |
| seq-length | 5 (default) | The length of continuous steps for LSTM training |

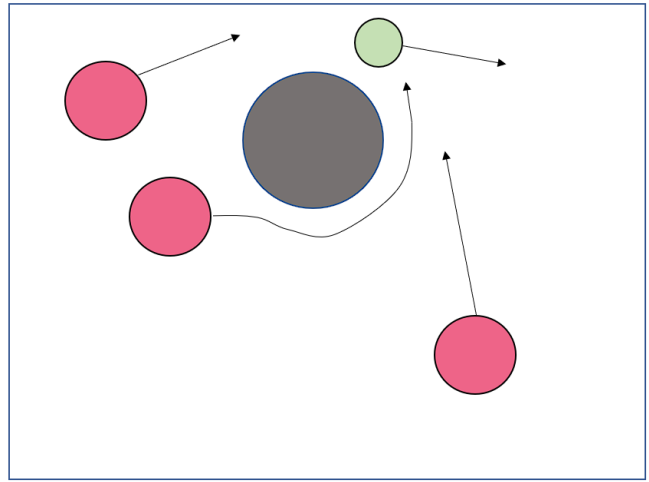

Figure 4. Illustration of Obstacle Predator Prey environment

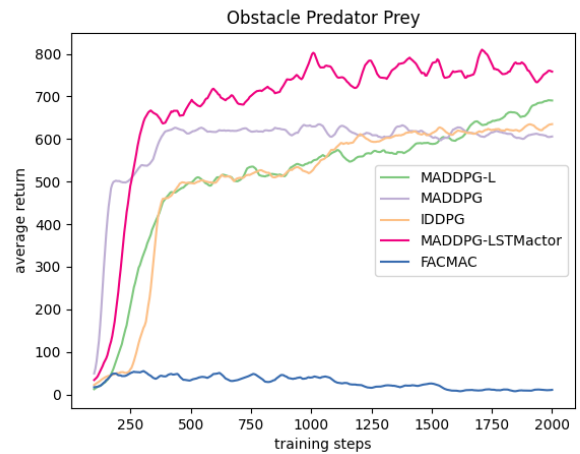

Figure 5. Algorithms' performance in Obstacle Predator Prey environment

The hyper-parameters used in the Experiment are listed in table 2, some of them remain the same in all environments and algorithms, others which vary from experiment to experiment will have their default value listed here and exact value shown in appendix.

**Obstacle Predator Prey**: In the Obstacle Predator-Prey environment, a team of agents will be trained to pursuit an adversarial agent with higher speed. The agents trained need to learn how to chase the target(prey) while avoiding crashing into obstacles and each other. The illustration of the environment is **Fig.** 4.

The result of the experiment is shown in **Fig.** 5. The result shows that algorithms using the CTDE training paradigm generally have better performance than IDDPG, which adopts the decentralized training method. FACMAC is the exception, the explanation can be found in chapter 3.2. In this environment, with the help of LSTM, MADDPG-LSTMactor has better performance than MADDPG. MADDPG-L, which only feeds agents' critics with the agent's own action and the environment state, demonstrated better performance than MADDPG in the final stage.

To find out the reason why FACMAC performed far worse than other algorithms, we did more experiments in this environment.

The first additional experiment is to test two different ways of getting the final centralized critic. As the calculation of it requires other agents' critic result besides the agent $a$ which is currently trained, we can directly use other agents' critic to estimate the action-value of their own actions, or use the critic network of agent a to simulate other agents' critic network to produce the action-value. As in our experiment, all agents' critic networks share the same structure, the simulation method is possible. The result of these two different calculation methods of the centralized critic in FACMAC is shown in **Fig.** 6.

It can be witnessed that using agent's critic network to simulate other agents' critic networks can reach a better result than directly using other agents' critic network, though the performance is still not comparable to other algorithms. The mixing network could be the reason. As in the FACMAC's design that the mixing network should be considered as a part of the critic network, the parameters of the mixing network will also be updated while optimizing the critic network, which brings uncertainty to the gradient's direction. This can be proved in our second additional experiment.

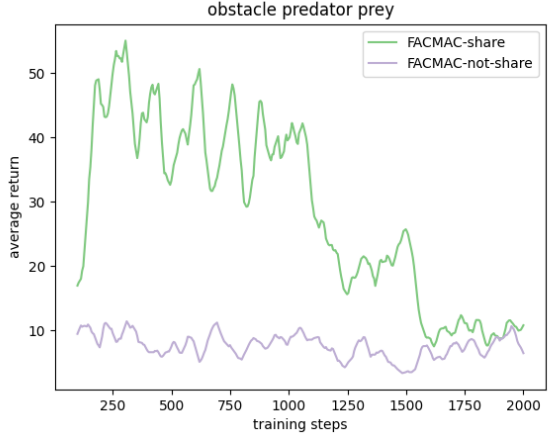

Figure 6.  Performance of FACMAC with different parameter sharing policy

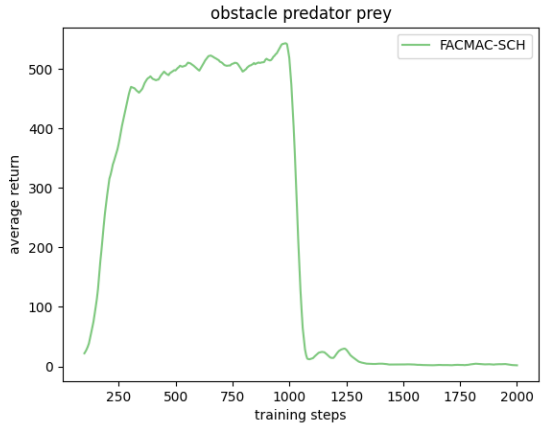

Figure 7.  FACMAC's performance by optimizing the critic networks and policy networks at the first stage, then training the mixer network

In our second additional experiment, we froze the parameters of the mixing network and only optimized agents' local critics in the first stage of training, thus the centralized critic was ablated. In the later stage, we started to update the mixing network's parameters and switched back to the normal training process of FACMAC. The way to determine the policy gradient also varied in two stages, Equation (13) demonstrates how it's done in the first stage and Equation (14) for the later stage.

$$\nabla_\theta J(\mu_a) = \mathbb{E}_\mathcal{D}\big[\nabla_\theta \mu_a \nabla_{\mu_a} Q_a^\mu(\tau_a, \mu_a(\tau_a); \phi_a)\big] \quad (13)$$

$$\nabla_\theta J(\boldsymbol{\mu}) = \mathbb{E}_\mathcal{D}\big[\nabla_\theta \boldsymbol{\mu} \nabla_{\boldsymbol{\mu}} \boldsymbol{Q}_{tot}^\mu(\tau, \mu_1(\tau_1), \mu_2(\tau_2) \ldots \mu_n(\tau_n), s)\big] \quad (14)$$

By training agents' local critic networks in advance, we aim to reduce the complexity of centralized training. Meanwhile, the successful experience tuples stored in the replay buffer assures that the algorithm can have enough positive data for training. The result of this experiment is shown in **Fig.** 7.

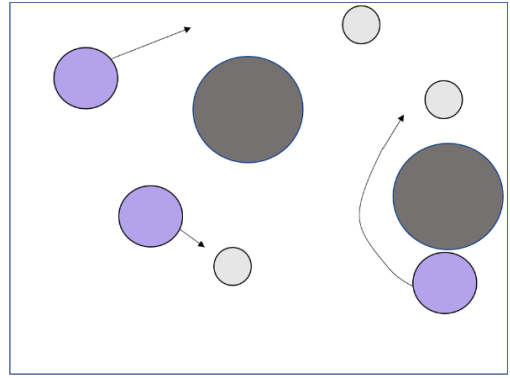

Figure 8.  The illustration of Spread environment

We set 1 million timestep to be the end of the first stage. The algorithm's performance simply plunged to the level of random policy after reaching the watershed. We altered each agent's local critic network from $Q_a^\mu(\tau_a, u_a; \theta_a)$ to $Q_a^\mu(s, u_a; \theta_a)$ and get rid of the mixing network. By doing this, we created MADDPG-L. The input of the algorithm's critic network doesn't have the problem of the explosion of action space as it only requires the agent's own action, and still provides global information.

**Spread**: In the Spread environment, target locations will be randomly generated after each environment reset, a team with the same amount of the target locations are trained to occupy all target locations. Obstacles with random location will also be generated in the space. Agents learn to occupy all target locations while avoid collision with the obstacle or each other. The illustration of this environment is **Fig.** 8.

Due to the different goal of training, the reward function in this environment is changed from the one in the Obstacle Predator-Prey. The reward function is set to be the negative distance sum of all target locations to their nearest agent. In other words, with each agent getting closer to the it's target, the reward will get higher. We also give agents negative rewards for having collisions with other agents or obstacles.

In order to examine algorithms' ability to scale to larger amount of agents, the Spread environment will have different versions with different amount of agents, Spread-3a with 3 agents, Spread-6a with 6 agents and Spread-9a with 9 agents. The result of algorithms in these 3 environments' performance are shown in **Fig.** 9.

From the result of experiments in those 3 environments, it can be seen that MADDPG-LSTMactor will perform better than MADDPG when the agents are relatively few. However, with the increase of agents, MADDPG-LSTMactor has no advantage over MADDPG. This could because the increased complexity brought by LSTM. Although MADDPG-L doesn't perform any better than MADDPG when the amount of agents are small, it demonstrates significant better performance than MADDPG as agents become more. It can be concluded that in this environment, MADDPG-L can scale to more agents better.

**Tunnel and Simple Tunnel**: In the Tunnel environment, two massive obstacles will form a narrow tunnel which forces the agents to plan path to go through. $N$ target locations spread at one end of the tunnel with fixed positions. After every environment reset, $N$ agents will emerge at fixed positions at the other end of tunnel. They need to go through tunnel and occupy every target location while avoid any contact with each other or the obstacle. The illustration of the environment is shown in **Fig.** 10(left).

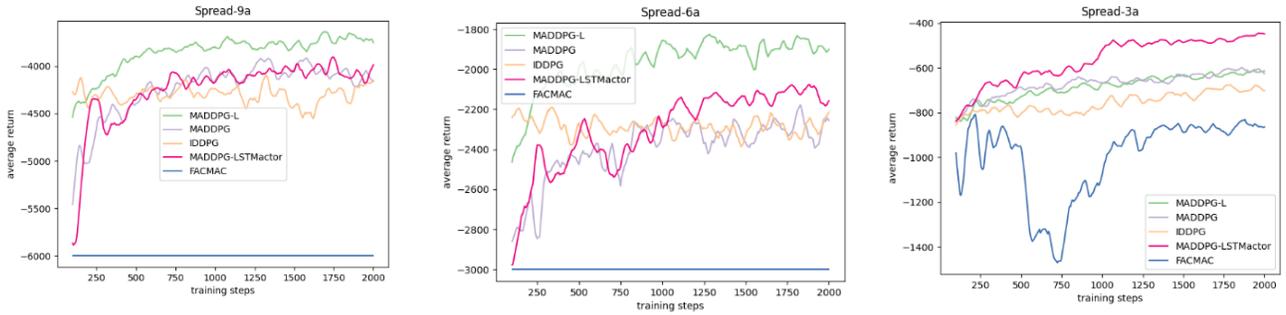

Figure 9. Algorithms' performance in Spread environment with different amount of agents.

Similar to the Spread environment, the reward function of the Tunnel environment is also set to be the negative sum of each target location to its nearest agent. Agents will get penalized for collide with other agents or the obstacle.

In order to simplify the environment as well as to be closer to the reality, some modifications are made to the Tunnel environment. We set walls around the obstacle to limit the agents' mobility range. The same amount of agents and target locations still emerge at two end of the tunnel. What's different is that each agent is designated one target location of its own to occupy. Agents also need to avoid collisions while moving to the target locations. The modified environment is named as the Simple Tunnel environment with its illustration shown in **Fig.** 10

The reward function is changed since agents in the Simple Tunnel environment have different goal. Their rewards are no longer collaborative and shared. The reward function of agent $a$ is determined by its distance to its designated target location.Penalties for collisions will also be given. Since each agent's reward function are not the same, FACMAC is not applicable in this environment.

The experiment results are shown in **Fig.** 11. From the test result, MADDPG-LSTMactor maintains its advantage over MADDPG when the number of agents is small. In the Simple Tunnel environment where is closer to reality, MADDPG-

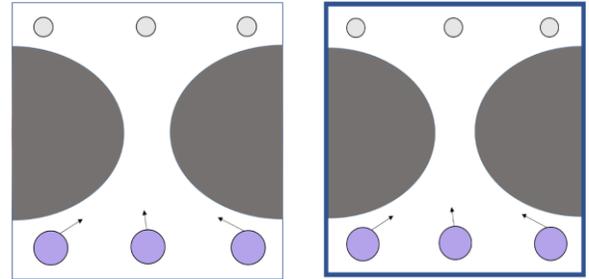

Figure 10. The illustration of Tunnel environment (left) and Simple Tunnel environment (right)

LSTMactor does not only demonstrate better performance, but also shows faster learning speed. While MADDPG-L's performance has no edge over MADDPG's, MADDPG-L also has a faster convergence speed than MADDPG.

After the experiment done in the Simple Tunnel environment, we increase the number of agents in the Simple Tunnel environment from 3 to 6 and name the new environment as Simple Tunnel 6a. The test result of algorithms in this environment is shown in **Fig.** 11(two on the right). The experiment demonstrated that MADDPG-L has a better performance than MADDPG with the increase of agents while MADDPG-LSTMactor lost its advantage over MADDPG in term of performance.

V. CONCLUSION

In this paper, we proposed two algorithms, viz., MADDPG-LSTMactor and MADDPG-L. These two algorithms and the underlying algorithms: IDDPG, MADDPG, FACMAC were together tested in several virtualenvironments to examine their ability to avoid obstacle and to cope with the increase in the number of agents.

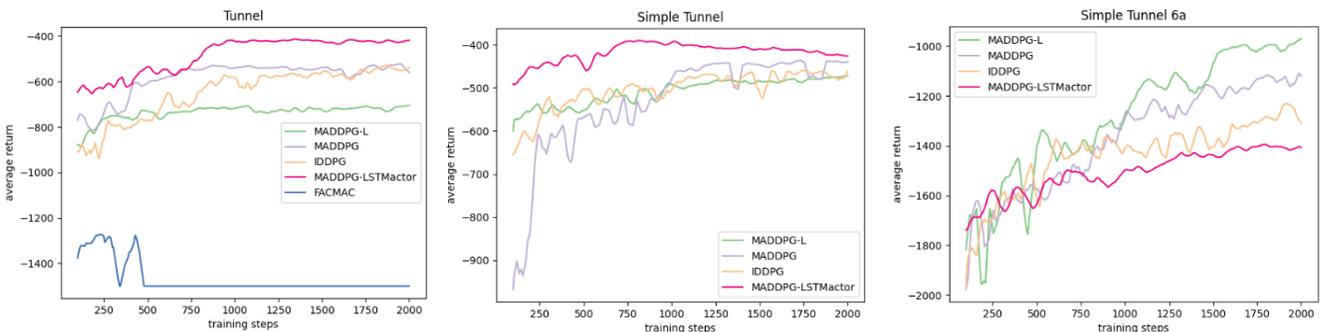

Figure 11. Algorithms' performance in Tunnel environmen(first on the left)) and Algorithms' performance in Simple Tunnel environment with different number of agents(two on the right)

After experiments, this paper can draw a preliminary conclusion: when the number of agents is small, the MADDPG-LSTMactor algorithm is a better multi-agent obstacle avoidance navigation algorithm than MADDPG. When the number of agents increases, the MADDPG-L algorithm demonstrates better performance.

The algorithms proposed in this paper exhibits certain capabilities in virtual environments, and can be used as a basic in the future. Of course, a realistic obstacle avoidance algorithm is by no means limited to using reinforcement learning as the only solution to a problem. There are also promising solutions of multi-agent navigation based on other frameworks, such as meta learning [24]. By combining geolocation, computer vision, etc., better performance may be achieved. Multi-agent navigation in case of more difficult scenarios, e.g., with limited communication power [25], is also well worth exploration.

APPENDIX

TABLE III. HYPERPARAMETERS IN OBSTACLE PREDATOR PREY ENVIRONMENT

| Hyperparameter | Value |
|---|---|
| Num-adversaries | 1 |
| Lr-actor | 0.001 |
| Lr-critic | 0.01 |
| Batch-size | 256 |
| seq-length | 3 |

TABLE IV. HYPERPARAMETERS IN SPREAD-3A ENVIRONMENT

| Hyperparameter | Value |
|---|---|
| Num-adversaries | 0 |
| Lr-actor | 0.001 |
| Lr-critic | 0.01 |
| Batch-size | 128 |
| seq-length | 5 |

TABLE V. HYPERPARAMETERS IN SPREAD-6A AND SPREAD-9A ENVIRONMENT

| Hyperparameter | Value |
|---|---|
| Num-adversaries | 0 |
| Lr-actor | 0.001 |
| Lr-critic | 0.01 |
| Batch-size | 32 |
| seq-length | 3 |

TABLE VI. HYPERPARAMETERS IN TUNNEL AND SIMPLE TUNNEL ENVIRONMENT

| Hyperparameter | Value |
|---|---|
| Num-adversaries | 0 |
| Lr-actor | 0.001 |
| Lr-critic | 0.01 |
| Batch-size | 128 |
| seq-length | 5 |

TABLE VII. HYPERPARAMETERS IN SIMPLE TUNNEL-6A ENVIRONMENT

| Hyperparameter | Value |
|---|---|
| Num-adversaries | 0 |
| Lr-actor | 0.001 |
| Lr-critic | 0.001 |
| Batch-size | 32 |
| seq-length | 3 |

ACKNOWLEDGMENT

This work was supported by the National Natural Science Foundation of China under grants 62172072, 61872101, 61872055, U1908214, U21A20491, the Fundamental Research Funds for the Central Universities under grant DUT21JC18, Natural Science Foundation of Liaoning Province of China under Grant 2021-MS-114.